\documentclass[11pt,a4paper]{article}
\usepackage[nohyperref]{emnlp2019/emnlp-ijcnlp-2019}
\usepackage{times}
\usepackage{latexsym}

\usepackage{url}            
\usepackage{booktabs}       
\usepackage{amsfonts}       
\usepackage{nicefrac}       
\usepackage{microtype}      

\usepackage{graphicx}
\usepackage{amsmath}
\usepackage{subcaption}
\usepackage{comment}

\usepackage{xcolor}

\aclfinalcopy 

\author{John P. Lalor$^{1}$\thanks{~~Current affiliation: Mendoza College of Business, University of Notre Dame}, Hao Wu$^{2}$, Hong Yu$^{1,3,4}$\\
$^1$College of Information and Computer Sciences, UMass Amherst \\
$^2$Department of Psychology and Human Development, Vanderbilt University\\ 
$^3$Department of Computer Science, UMass Lowell\\
$^4$Center for Healthcare Organization \& Implementation Research, Bedford VA\\
\tt{john.lalor@nd.edu}, \tt{hao.wu.1@vanderbilt.edu}, \tt{hong.yu@umassmed.edu}}

\title{Learning Latent Parameters without Human Response Patterns: \\Item Response Theory with Artificial Crowds}

\date{}

\begin{document}
	\maketitle

\begin{abstract}
	Incorporating Item Response Theory (IRT) into NLP tasks can provide valuable information about model performance and behavior.
	Traditionally, IRT models are learned using human response pattern (RP) data, presenting a significant bottleneck for large data sets like those required for training deep neural networks (DNNs).
	In this work we propose learning IRT models using RPs generated from artificial crowds of DNN models.
	We demonstrate the effectiveness of learning IRT models using DNN-generated data through quantitative and qualitative analyses for two NLP tasks.
	Parameters learned from human and machine RPs for natural language inference and sentiment analysis exhibit medium to large positive correlations.
	We demonstrate a use-case for latent difficulty item parameters, namely training set filtering, and show that using difficulty to sample training data outperforms baseline methods.
	Finally, we highlight cases where human expectation about item difficulty does not match difficulty as estimated from the machine RPs.
\end{abstract}

\section{Introduction}

What is the most difficult example in the Stanford Natural Language Inference (SNLI) data set \cite{bowman_large_2015} or
in the Stanford Sentiment Treebank (SSTB) \cite{socher2013recursive}?
\textit{A priori} the answer is not clear.
How does one quantify the difficulty of an example 
and does it pertain to a specific model, or more generally?

There has been much recent work trying to assess the quality of data sets used for NLP tasks, \cite[e.g.][]{lalor2016beyond,sakaguchi2018efficient,kaushik2018much}. 
In particular, a common finding is that different examples within the same class have very different qualities such as difficulty, and
these differences affect models' performance.
For example, one study found that a subset of reading comprehension questions were so difficult as to be unanswerable \cite{kaushik2018much}.
In another work, the difficulty of specific items was found to be a significant predictor of whether a model would classify the item correctly \cite{lalor2017analysis}.

While a number of methods exist for estimating difficulty, in this work we focus on Item Response Theory (IRT) \cite{baker2001basics,baker_item_2004}, a widely used method in psychometrics.
IRT models fit parameters of data points (called ``items'') such as difficulty based on a large number of annotations (``response patterns'' or RPs), typically gathered from a human population (``subjects''). 
It has been shown to be an effective way to evaluate and analyze NLP models with respect to human populations \cite{lalor2016beyond,lalor2017analysis}.

While IRT models are designed to be learned with human RPs for at most $100$ items, data sets used in machine learning, particularly for training deep neural networks (DNNs), are on the order of tens or hundreds of thousands of examples or more.
It is not possible to ask humans to label every example in  a data set of that size.
In this work we hypothesize that IRT models can be fit using RPs from artificial crowds of DNNs as inputs, thereby removing the expense of gathering human RPs.
Recent work has shown that DNNs encode linguistic knowledge \cite{tenney2019you,tenney2019bert} and can reach or surpass human-level performance on classification tasks \cite{lake2015human}.  
In addition, generating IRT data with deep learning models is much cheaper compared to employing human annotators.

We demonstrate that learned parameters from IRT models fit with artificial crowd data are positively correlated with parameters learned with human data for small data sets.
We then use variational inference (VI) methods \cite{jordan1999introduction,hoffman2013stochastic} to fit a large-scale IRT model.
Using VI allows us to scale IRT models to deep-learning-sized data sets.
Finally, we show why learning such models is useful by demonstrating how learned difficulties can improve training set subsampling. 

Our contributions are as follows: (1) We show that IRT models can be fit using machine RPs by comparing item parameters learned from human and from machine RPs for two NLP tasks; (2) we show that RPs from more complex models lead to higher correlations between parameters from human and machine RPs; (3) we demonstrate a use-case for latent difficulty item parameters, namely training set filtering, and show that using difficulty to sample training data outperforms baseline methods; (4) we provide a qualitative analysis of items with the largest human-machine disagreement in terms of difficulty to highlight cases where human intuition is inconsistent with model behavior.

These results provide a direct comparison between humans and machine learning models in terms of identifying easy and difficult items. 
They also provide a foundation for large-scale IRT models to be fit by using ensembles of machine learning models to obtain RPs instead of humans, greatly reducing the cost of data-collection.\footnote{Code for IRT model fitting is available at \url{https://github.com/jplalor/py-irt}.}

\section{Fitting Item Response Theory Models}
\label{sec:irt}
\subsection{Traditional Item Response Theory}

Here we briefly describe IRT and the specific model under consideration, the Rasch model (also known as the one-parameter logistic or 1PL model) \cite{rasch1960studies}.

We refer the reader to \cite{baker2001basics,baker_item_2004} for additional details on IRT, and to \cite{martinez2016making,lalor2016beyond,lalor2017analysis} for more details on previous applications of IRT to machine learning.

IRT models are designed to estimate latent ability parameters ($\theta$) of subjects and latent item parameters such as difficulty of items ($b$).
For a 1PL model, the probability that subject $j$ will answer item $i$ correctly is a function of the subject's latent ability $\theta_j$ and the item's latent difficulty $b_i$:
\begin{equation} 
\label{eq:irt}
p(y_{ij} = 1 \vert \theta_j, b_i) = \frac{1}{1 + e^{-(\theta_j - b_i)}}
\end{equation}
The probability that subject $j$ will answer item $i$ incorrectly is: 
\begin{equation} 
p(y_{ij} = 0 | \theta_j, b_i) = 1 - p(y_{ij} = 1 \vert \theta_j, b_i) 
\end{equation}
The likelihood of a data set of RPs $Y$ from $J$ subjects to a set of $I$ items is:
\begin{align} 
p(Y \vert \theta, b) &= \prod_{j=1}^J \prod_{i=1}^I p(Y_{ij}=y_{ij} \vert \theta_j, b_i)
\end{align} 

For the 1PL model, the difficulty parameter represents the ability level at which the probability of an individual answering an item correctly is 50\%.
This occurs when item difficulty is equal to subject ability ($\theta_j = b_i$ in Eq. \ref{eq:irt}).

The item parameters are typically estimated by marginal maximum likelihood (MML) via an Expectation-Maximization (EM) algorithm \cite{bock1981marginal}, in which subject parameters are considered random effects $\theta_i\sim N(0,\sigma_\theta^2)$ and marginalized out.
Once item parameters are learned, subjects' $\theta$ parameters are scored typically with maximum a posteriori (MAP) estimation.
IRT models are usually fitted to RPs of hundreds or thousands of human subjects, who usually answer at most $100$ questions.
Therefore the methods for fitting these models have not been scaled to huge data sets and large numbers of subjects (e.g. tens of thousands of machine learning models).

\subsection{IRT with Variational Inference}

VI is a model fitting method that approximates an intractable posterior distribution in Bayesian inference by a simpler variational distribution.
Prior work has compared VI methods with traditional IRT methods \cite{natesan2016bayesian} and found it effective, but was primarily concerned with fitting IRT models for human-scale data.

Bayesian methods in IRT assume that the individual $\theta$ and $b$ parameters in Eq. (2) both follow Gaussian prior distributions and make inference through the resultant joint posterior distribution $\pi(\theta,b|Y)$. As this posterior is usually intractable, VI approximates it by the variational distribution:
\begin{align} 
q(\theta, b) &=  \prod_{j=1}^J \pi^\theta_j(\theta_j) \prod_{i=1}^I \pi^b_i(b_i)
\end{align} 

Where $\pi^\theta_j()$ and $\pi^b_i()$ denotes different Gaussian densities for different parameters whose means and variances are determined by minimizing the KL-Divergence between $q(\theta,b)$ and $\pi(\theta,b|Y)$.

The choice of priors in Bayesian IRT can vary.
Prior work has shown that vague and hierarchical priors are both effective \cite{natesan2016bayesian}.
We experiment with both in this work. A vague prior assumes $\theta_j \sim N(0, 1)$ and $b_i \sim N(0, 10^3)$, where the large variance indicates a lack of information on the difficulty parameters.
A hierarchical Bayesian model assumes
\begin{align*}
\theta_j\ |\ m_\theta, u_\theta &\sim N(m_{\theta}, u^{-1}_{\theta}) \\
b_i\ |\ m_b, u_b &\sim N(m_b, u^{-1}_b) \\
m_{\theta}, m_{b} &\sim N(0, 10^6) \\
u_{\theta}, u_b &\sim \Gamma(1, 1)
\end{align*}
Our results for these two options were very similar, so we only report those for hierarchical priors.
\section{Data and Models}

Here we describe the data sets used to conduct our experiments, as well as the DNN model architectures for both generating response patterns and conducting our training set filtering experiment.

\paragraph{SNLI}
The SNLI data set \cite{bowman_large_2015} is a popular data set for the natural language inference task.
Briefly, each example in the data set consists of two sentences in English: the premise and the hypothesis, and a corresponding label.
The correct label is ``entailment'' if the premise implies the hypothesis, ``contradiction'' if the premise implies that the hypothesis must be false, and ``neutral'' if the premise implies neither the hypothesis nor its negation.
SNLI consists of 550k/10k/10k training/validation/testing examples examples.

\paragraph{SSTB} 
The Stanford Sentiment Treebank (SSTB) \cite{socher2013recursive} is a collection of English phrases extracted from movie reviews with fine-grained sentiment annotations (very negative, negative, neutral, positive, very positive).
In this work we focus on binary sentiment classification, using the SST-2 split of the data set, where neutral examples have been removed. 
The data set consists of 67k/873/1.8k training/validation/testing examples.

\paragraph{Human RP Data}
The human RP data sets for SNLI and SSTB were previously collected  from Amazon Mechanical Turk (AMT) workers \cite{lalor2016beyond,lalor2017analysis}.
For a randomly selected sample of items from SNLI and SSTB, new labels were gathered from 1000 AMT workers (Turkers). 
Each Turker labeled each item, so that for each item there were 1000 new labels.
For each Turker, a RP was generated by grading the provided labels against the known gold-standard label.

\paragraph{Building an Artificial Crowd}

As mentioned earlier, it is not feasible to have humans provide RPs for data sets used to train DNN models.
Can we instead use RPs from DNNs?
We trained an ensemble of DNN models with varying amount of training data to simulate an artificial crowd so that enough responses were obtained to fit the IRT models.
The goal here is not to build an ensemble of DNNs to surpass current classification state of the art results, but instead to test our hypothesis to determine if machine RPs can fit IRT models that can benefit NLP tasks.

Specifically, we trained $1000$ LSTM models for NLI classification using the SNLI data set and $1000$ LSTM models for binary SA classification using the SSTB data set \cite{bowman_large_2015, socher2013recursive}.
The SNLI model consists of two LSTM sequence-embedding models \cite{hochreiter_long_1997}, one to encode the premise and another to encode the hypothesis.
The two sentence encodings are then concatenated and passed through three tanh layers.
Finally, the output is passed to a softmax classifier layer to output class probabilities.
For SSTB, we used a single LSTM model without the concatenation step.
The models were implemented in DyNet \cite{dynet}.
Models were trained with SGD for 100 epochs with a learning rate of 0.1, and validation set accuracy was used for early stopping.

For each model $m_i$, we randomly sampled a subset of the task training set, $x_{\text{train}}^i$.
We corrupted a random selection of training labels by replacing the gold standard label with an incorrect label.
For each model-training set pair, we trained the model, used the held out validation set for early stopping, and wrote the model's graded (correct/incorrect) outputs to disk as that model's RP.
The set of RPs for all models is our input data for the IRT models.

We also looked at a more complex model to determine if the learned parameters would differ given the different model architectures.
For our more complex model we used the Neural Semantic Encoder model (NSE), a memory-augmented recurrent neural network \cite{munkhdalai2016neural}:
\allowdisplaybreaks
\begin{align*}
o_t &= f_{\text{r}}^{LSTM}(x^t) \\
z_t &= softmax(o_t^{\top}M_{t-1}) \\
m_{r,t} &= z_t^{\top}M_{t-1} \\
c_t &= f_{\text{c}}^{MLP}(o_t,m_{r,t}) \\
h_t &= f_{\text{w}}^{LSTM}(c_t) \\
\begin{split}
M_t = M_{t-1}(\mathbf{1} &- (z_t \otimes e_k)^{\top}) \\
&+ (h_t \otimes e_t)(z_t \otimes e_k)^{\top}
\end{split}
\end{align*}
where $f_r^{LSTM}$ is the read function, $f_c^{MLP}$ is the composition function, $f_w^{LSTM}$ is the write function, $M_t$ is the external memory at time $t$, and $e_l \in R^l$ and $e_k \in R^k$ are vectors of ones.

The goal with the data set restriction and label corruption was to build an ensemble of models with widely varying performance on the SNLI test set.
Training with different training set sizes and levels of noise corruption means that certain models will perform very well on the test set (large training sets and low label corruption) while others will perform poorly (small training sets and high label corruption).
This way we will get a variety of response patterns to simulate performance on the task across a spectrum of ability levels.
While we could have modified the networks in any number of ways (e.g. changing layer sizes, learning rates, etc.), modifying the training data is a straightforward method for generating a variety of response patterns, and has been shown to have an impact on performance in terms of item difficulty \cite{lalor2017analysis}.
Further investigations of network modifications is left for future work.

\section{Methods}
We conduct the following experiments: (i) a comparison of IRT parameters learned from human and machine RP data, using existing IRT data sets \cite{lalor2016beyond,lalor2017analysis} as the baseline for comparison, (ii) a comparison between MML and VI parameter estimates, and (iii) a demonstration of the effectiveness of learned IRT parameters via training data set selection experiments.

\subsection{Validating Variational Inference}

Before using VI to fit IRT models for DNN data, we must first show that VI produces estimates similar to traditional methods.
This was established in prior work on synthetic data \cite{natesan2016bayesian}.
Here we compare them on an existing human data set \cite{lalor2016beyond}.

A traditional Rasch model was fit with both MML and VI. MML was implemented in the R package mirt \cite{chalmers_mirt:_2015} and VI in Pyro \cite{bingham2018pyro}, a probabilistic programming language built on PyTorch \cite{paszke2017automatic} that implements typical VI model fitting and variance reduction \cite{kingma2013auto,ranganath2014black}.
We calculate the root mean squared difference (RMSD) between MML and VI estimates for subject and item parameters.
Our expectation is that the RMSD will be sufficiently small to confirm that the VI parameters are similar enough to those learned by MML, since we will not be able to use MML when we attempt to scale up to larger data sets. 

\begin{figure*}[th!]
	\centering
	\begin{subfigure}[b]{0.49\textwidth}
		\centering
		\includegraphics[width=0.95\columnwidth]{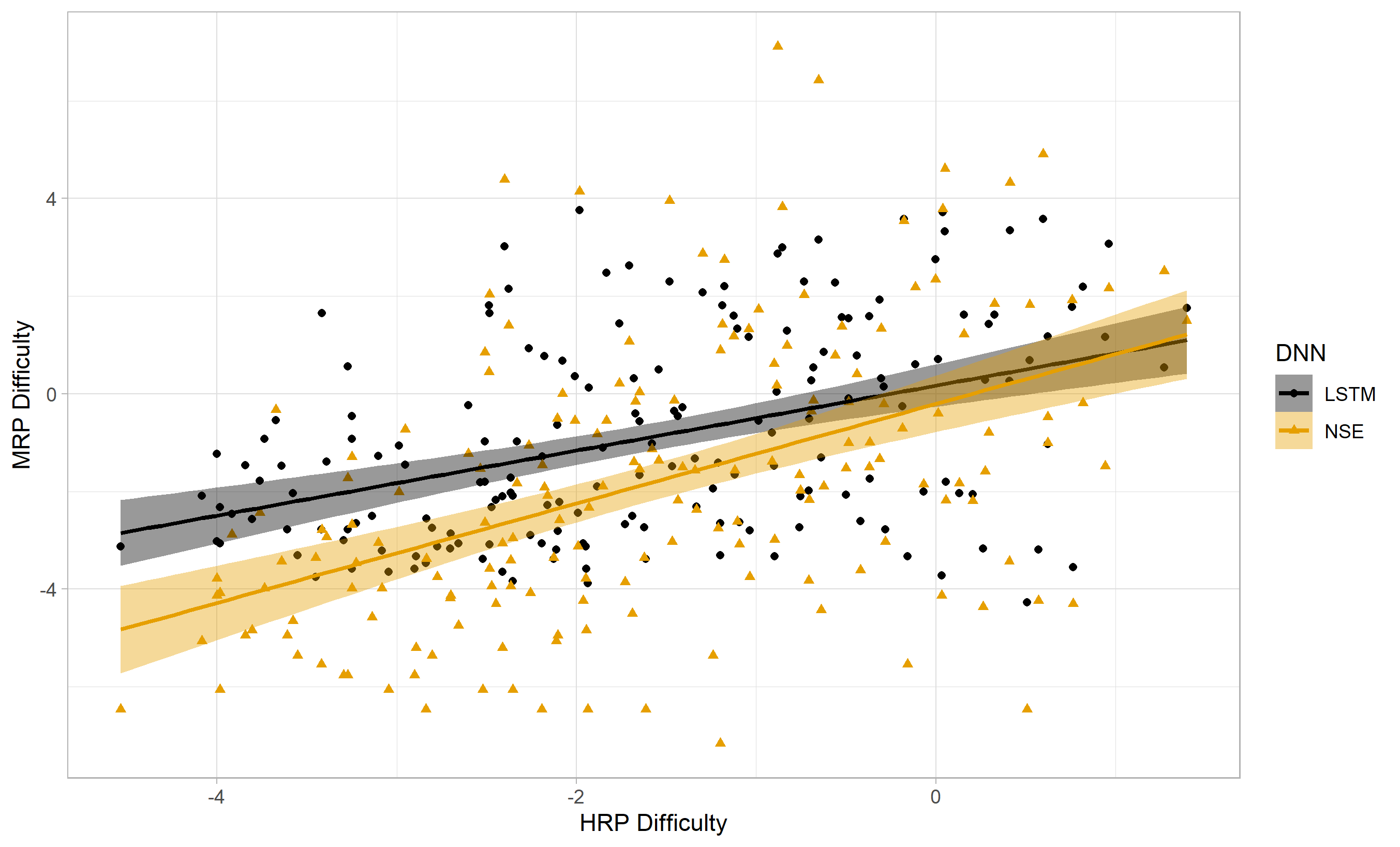}
		\caption{NLI}
		\label{fig:diffs_nli}
	\end{subfigure} 
	\begin{subfigure}[b]{0.49\textwidth}
		\centering
		\includegraphics[width=0.95\columnwidth]{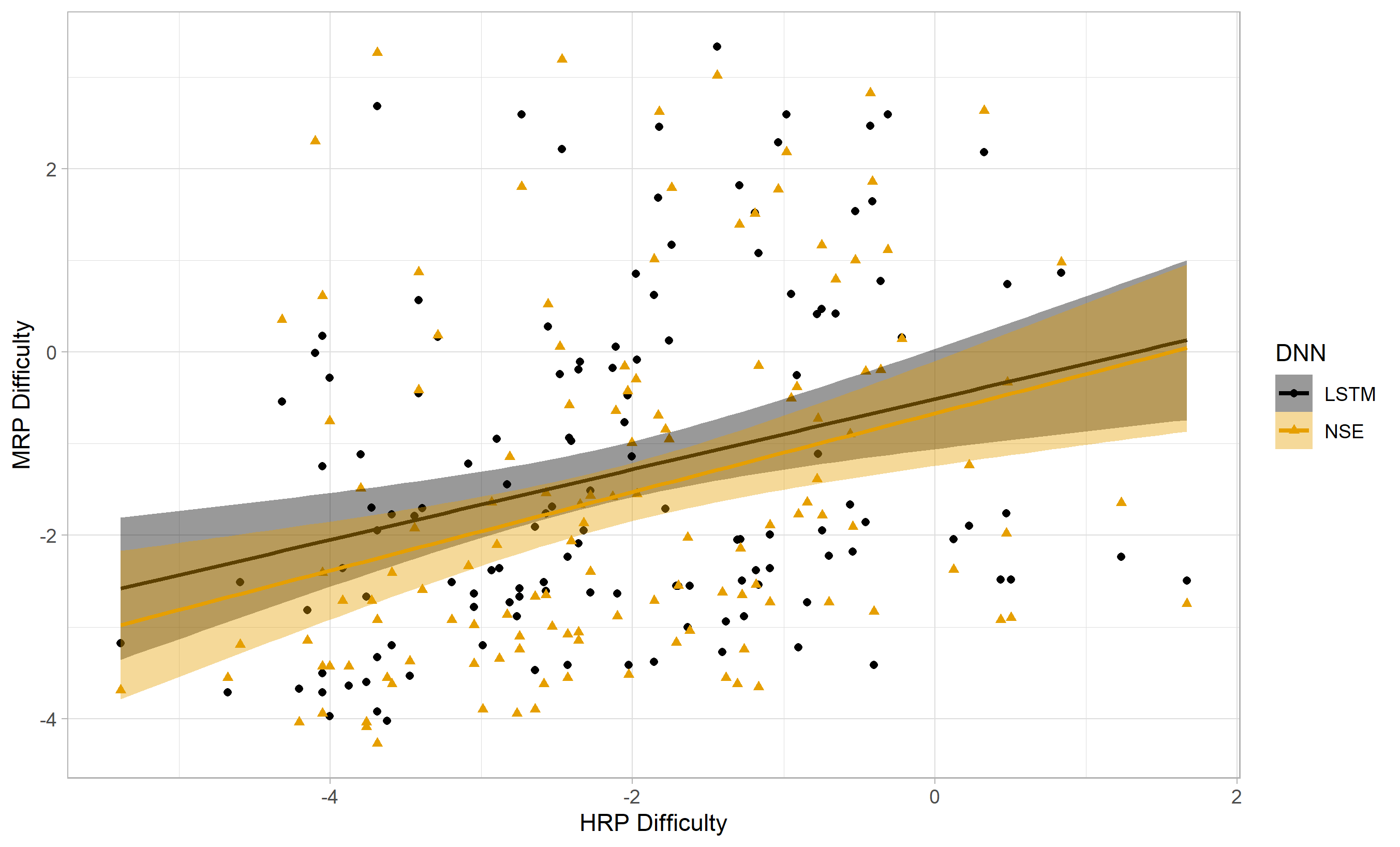}
		\caption{SA}
		\label{fig:diffs_sa}
	\end{subfigure} 
	
	\caption{Comparison of learned item difficulty parameters for human (x-axis) and machine data (y-axis) for NLI (Fig. \ref{fig:diffs_nli}) and SA (Fig. \ref{fig:diffs_sa}). Spearman $\rho$ (NLI): $0.409$ (LSTM) and $0.496$ (NSE). Spearman $\rho$ (SA): $0.332$ (LSTM) and $0.392$ (NSE).}
	\label{fig:diffs_nli_sstb}
\end{figure*}

\subsection{Human Machine Correlation}

We further compare item difficulty parameters learned from machine RPs to those learned from human RPs. 
These two sets of parameters cannot be compared directly as they can only be interpreted in reference to their respective subject populations.
Instead, we compute the correlation between these two sets of parameters to see whether items that are easy for humans are also easy for machines.
We fit two Rasch models, one with existing human RPs \cite{lalor2016beyond,lalor2017analysis}. and one with the machine RPs.
Both models were fit with MML using the mirt R package \cite{chalmers_mirt:_2015}.
Learned item difficulty parameters were extracted and compared via Spearman $\rho$ rank order correlations.

\subsection{Training Set Subsampling}

To demonstrate the usefulness of the learned IRT parameters, we next describe a downstream use case: training set filtering for more efficient learning.
Can we maintain model performance by removing the easiest and/or hardest items from the training set?
Once difficulty parameters for each data set were learned, we trained a new DNN model using only a subset of the original training data.
We trained a number of models, each with a different cutoff in terms of training data to observe how generalization was impacted in each case.

We looked at 4 filtering strategies (in each case $d$ is the item difficulty threshold): (i) absolute value inner (AVI), where all training items with $|b_i| < d$ were retained, (ii) absolute value outer (AVO), where all training items with $|b_i| > d$ were retained, (iii) an upper bound (UB), where items with $b_i < d$ were retained, and (iv) a lower bound (LB), where items with $b_i > d$ were retained.
These methods were compared against two baselines that consider the percentage of models that label an item correctly ($0 \leq pc \leq 1$) as an inexpensive proxy for difficulty: (i) percent-correct upper bound (PCUB), where items with $pc_i < d$ were retained, and (ii) percent-correct lower bound (PCLB), where items with $pc_i > d$ were retained.
Setting an upper bound on difficulty (UB) is similar to setting a lower bound on percent correct (PCLB) (i.e., we are excluding the hardest items from training).
Similarly, setting a lower bound on difficulty (LB) is analogous to setting an upper bound on percent correct (PCUB) in that they both exclude the easiest items from training.

Each of the filtering strategies have arguments in favor of their potential effectiveness.
AVI includes ``average'' items in terms of training examples, none that are too easy or too difficulty.
AVO is the opposite, where only the easiest and most difficult examples are retained, so that the extremes for each class can be learned.
UB ensures that those examples that are too difficult are not included, and LB ensures that the examples that are too easy are not included so that the model doesn't spend time learning very easy examples.

\section{Results}
\subsection{Human Machine Model Correlations}
\label{ssec:correlations}

\begin{figure*}[th!]
	\centering
	\includegraphics[width=2\columnwidth]{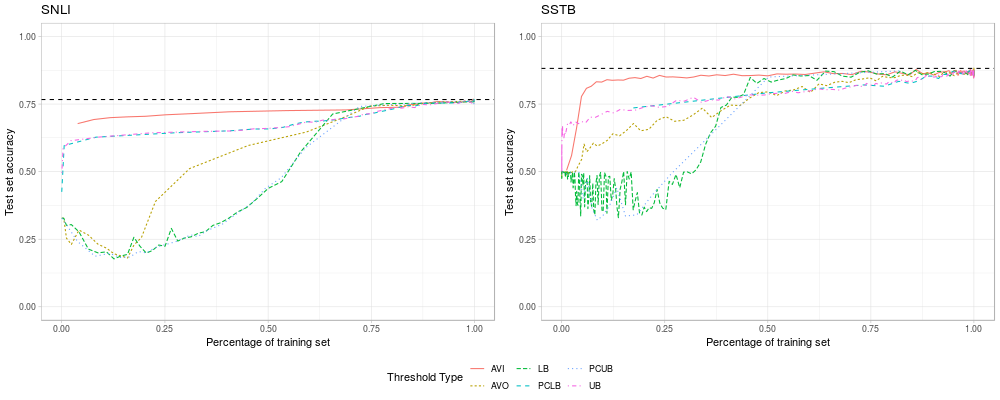}
	\caption{Test set accuracy by filtering strategy for NLI (left) and SA (right) plotted against percentage of training data retained. In both tasks filtering using the AVI strategy is most efficient in terms of high accuracy for small training set sizes.}
	\label{fig:thresholds}
\end{figure*}

We first look at the results of our human-machine model comparison (Figures \ref{fig:diffs_nli} and \ref{fig:diffs_sa}).
As an upper bound for correlations, we split the human annotation data in half for both SNLI and SSTB, fit two IRT Rasch models, and calculated the correlation between the learned parameters.
Spearman $\rho$ values were 0.992 and 0.987 for SNLI and SSTB items, respectively.

For both SNLI and SSTB, we find a positive correlation between the item difficulties of IRT models fit using human and machine RPs.
In addition, the more complex NSE model has consistently a higher correlation with the human-learned difficulty parameters than the LSTM model.
This suggests that creating more complex DNN architectures has bearing on how the model identifies difficult items with regards to human expectations.

The correlation is not perfect, and we would argue that this is an expected and encouraging result.
A close to perfect correlation would indicate that the DNN models and the human population agree closely on the difficulty ranking for the data sets and would be an incredible finding and evidence for the argument that DNN models encode human knowledge well, at least with respect to the difficulty of specific items.
This of course is not true, and the positive but not perfect correlation coefficients indicate this as such.
That said, it is encouraging that the positive correlation exists.
One would expect that training ensembles of more sophisticated NLP models such as BERT \cite{devlin2018bert} would further increase correlation scores. 



\subsection{Learning IRT Models with VI}
\label{ssec:irtvi}
Our next goal was to determine if VI could be used to fit IRT models and confirm prior work to that effect \cite{natesan2016bayesian}.
The RMSDs between MML and VI estimates were $0.158$ and $0.154$, respectively, for the difficulty and ability parameters.
Learned parameters are very similar between the two methods, which is to be expected.
This echos the results of prior work showing that VI is a good alternative to traditional MML methods for learning IRT models \cite{natesan2016bayesian}.
This result holds not only with synthetic data, as was used in the prior work, but also with human data collected for the development of an actual IRT test \cite{lalor2016beyond}.

\subsection{Data Filtering}

\begin{figure}[th!]
	\centering
	\includegraphics[width=0.8\columnwidth]{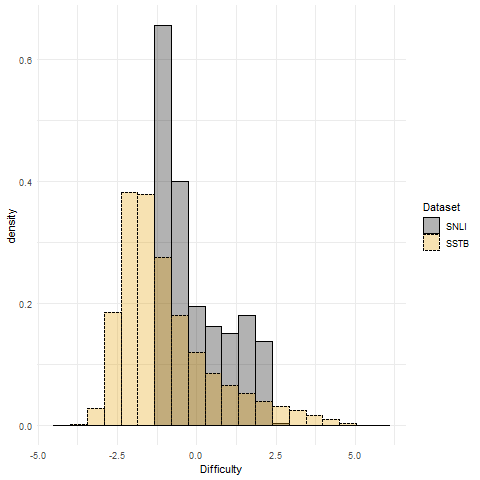}	
	\caption{Density plot of learned difficulties for SNLI and SSTB data sets.}
	\label{fig:merged_hists}
\end{figure}

Finally we consider training new DNN models on the filtered training data sets, restricted according to latent difficulty and the strategies described above (Figure \ref{fig:thresholds}).
The horizontal dotted lines in each plot represent the test set accuracy for a model trained with the full training data set.
For both SNLI and SSTB, the AVI strategy of selecting ``average'' examples leads to very good test set accuracy scores with less than 25\% of the original training data.
This shows that the strategy of selecting training data in terms of average difficulty, and gradually adding easier and harder examples at the same time provides examples that allows trained models to generalize well.
For both tasks, there is a large number of examples that are very easy in terms of latent difficulty (Figure \ref{fig:merged_hists}).
Sampling with AVI avoids selecting too many examples that are too easy and instead selects examples that are of average difficulty for the task, which may be better for learning.
In both cases LB and PCUB are the least effective strategies, indicating that it is not enough to only include the most difficult examples. 

The plots show that PCUB and LB provide very similar results, as do PCLB and UB, which is to be expected.
Difficulty parameters learned from IRT are very similar to metrics such as percent correct, but as the plots show are not exactly the same. 
Differences in RPs (i.e. which specific items were answered correctly/incorrectly) have an effect on item difficulty that is not captured by calculating percent correct.

It is worth noting here that the filtering strategy we used did not take class labels into consideration.\footnote{This is true for only the filtering step. Class labels are needed for learning the difficulty parameters needed for filtering (\S \ref{sec:irt}).}
The only determining factor as to whether a training item was included was the learned difficulty parameter $b_i$, which led to class imbalances in the training set.
This imbalance, however did not seem to have a significant negative effect in terms of performance.
More advanced sampling strategies that maintain training set distribution or sample data using a Bayesian approach are left for future work.

\begin{table} 
	\centering 
	\small 
	\begin{tabular} {cccc}
		\toprule
		\bf Strategy  & \multicolumn{3}{c}{\bf \% of Training Data} \\ 
		&0.1\% & 1\% & 10\%  \\
		\cmidrule{2-4}
		Random (reported) & 82.1 & 85.2 & \bf 88.4 \\
		Random (small batch) & 81.79 & 84.90  & 88.32  \\ 
		Lower-bound &43.68  & 41.56 & 39.89 \\
		Upper-bound & 81.62 & 80.46 & 79.06 \\ 
		AVI & \bf 82.44 & \bf 85.44 & 86.73 \\
		AVO & 43.60 &42.05  & 40.81  \\
		\bottomrule 
	\end{tabular}
	\caption{Dev accuracy results for MT-DNN model with different training set sampling strategies.}
	\label{tab:mtdnn} 
\end{table} 

\begin{table*}[th!]
	\centering 
	\small
	\begin{tabular}{p{7.3cm}p{4.3cm}lc}
		\toprule 
		Premise &Hypothesis & \multicolumn{2}{c}{Label \quad\quad Difficulty }\\
		\midrule
		Two men and a woman are inspecting the front tire of a bicycle.&There are a group of people near a bike.&Entailment& -3.7\\
		A girl in a newspaper hat with a bow is unwrapping an item.&The girl is going to find out what is under the wrapping paper.&Entailment& 3.1\\
		\midrule
		Two dogs playing in snow.&A cat sleeps on floor&Contradiction& -4.0\\
		Man sweeping trash outside a large statue.&A man is on vacation.&Contradiction& 3.8\\
		\midrule
		People sitting in chairs with a row flags hanging over them.&A family reunion for Fourth of July&Neutral& -3.6\\
		A group of dancers are performing.&The audience is silent.&Neutral&3.8\\
		\bottomrule
	\end{tabular}
	\caption{The easiest and hardest items judged by machine responses for each class in the SNLI test data set.}
	\label{tab:rankings}
\end{table*}

\begin{table*}[t!]
	\centering 
	\small
	\begin{tabular}{llp{8.5cm}ccc}
		\toprule 
		Task & Label &Item Text & \multicolumn{3}{c}{Difficulty ranking} \\
		& && Humans & LSTM & NSE \\
		\midrule
		SNLI&Contradiction&\textit{P:} Two dogs playing in snow.\newline \textit{H:} A cat sleeps on floor & 168 & 1 & 5 \\
		\cmidrule{3-6}
		&Entailment&\textit{P:} A girl in a newspaper hat with a bow is unwrapping an item. \newline \textit{H:} The girl is going to find out what is under the wrapping paper. & 55 & 172 & 176 \\
		\cmidrule{1-6}
		SSTB &Positive&Only two words will tell you what you know when deciding to see it: Anthony. Hopkins.& 9 & 103 & 110 \\
		\cmidrule{3-6}
		&Negative& ...are of course stultifyingly contrived and too stylized by half. Still, it gets the job done--a sleepy afternoon rental. & 128 & 46 & 41 \\
		\bottomrule
	\end{tabular}
	\caption{Examples from the SNLI and SSTB data sets where the ranking in terms of difficulty varies widely between human and DNN models. In all cases difficulty is ranked from easy to hard (1=easiest).}
	\label{tab:diff_rankings}
\end{table*}

As an additional experiment, we used the learned difficulty parameters to compare data sampling strategies for a state-of-the-art NLI model, MT-DNN \cite{liu2019mt-dnn}.
We sampled training data for SNLI at several intervals (0.1\%, 1\%, 10\%) and trained the MT-DNN model with the sampled data. 
We trained each model, as well as the random sample baseline, using the publicly available MT-DNN code.\footnote{\url{https://github.com/namisan/mt-dnn}}
Results are reported in Table \ref{tab:mtdnn}.
Note that we report two random baselines: (i) those reported in the original work, which were obtained by training the MT-DNN model with a batch size of 32.
Due to GPU resource constraints we had to train each MT-DNN model with a batch size of 8, and therefore report our reproduced random baseline results that we obtained as well (``Random (small batch)'').
For very small samples of data, the AVI strategy outperforms random sampling and all other methods as well.
As more data is sampled, the random models perform better. 
This indicates that a more advanced sampling strategy that starts with AVI then incorporates outliers (very easy/hard examples) at certain thresholds may improve learning as well.

\section{Analysis} 

\paragraph{Qualitative Evaluation of Difficulty}

Table \ref{tab:rankings} shows examples of premise-hypothesis sentence pairs from SNLI with the learned difficulty parameter from the machine RP IRT model.
The easy sentence pairs for each class seem to be very obvious, whereas the most difficult examples are difficult due to ambiguity.
For example, the hardest contradiction example could be classified as neutral instead of contradiction. 
It could be the case that the man is sweeping while on vacation, though it isn't likely.
The hypothesis doesn't directly contradict the premise like the easy example does (cats instead of dogs, sleeping instead of playing).

\paragraph{Analysis of Differences}

An interesting question comes up as a result of the less-than-perfect correlation scores (\S \ref{ssec:correlations}): Where are the differences?
To examine these more closely we identified those examples from the data sets where the rank order was most different between the human- and machine-response pattern models (Table \ref{tab:diff_rankings}).
That is, we calculated the absolute difference in ranking between the human model and the DNN model, and selected those where that value was highest.
The average absolute difference in ranking was around 40 for the SNLI task and around 30 for SSTB, for both the LSTM and NSE ensembles.

We can see interesting patterns in the discrepancies.
For SNLI, the easiest sentence pair for the LSTM model (which is also very easy for the NSE model) is one of the hardest for humans (Table \ref{tab:diff_rankings}, row 1).
Upon inspection of the gathered labels, the high difficulty comes from the fact that there were many Turkers who labeled the data as neutral and also many who labeled it as contradiction.

On the other hand, an example that is easy for humans but difficult for the DNN models (Table \ref{tab:rankings}, row 2) requires more abstract thinking than the earlier example.
The humans are able to infer that because the girl is unwrapping an item, she will discover what is under the wrapping paper when the unwrapping is complete.
The models find this pair to be one of the most difficult in the data set.

For SSTB, we see similar patterns (Table \ref{tab:diff_rankings}, rows 3-4).
For humans, one of the easiest review snippets is clearly positive (row 3), mainly because we know who Anthony Hopkins is and know how to rate his quality as an actor.
However for the DNN models, the text itself does not have a lot of positive or negative signal and therefore the item is considered very difficult.
On the other hand, the last example is very difficult for humans (row 4), possibly due to the relatively neutral text.
However, for the DNN models certain terms such as ``stultifyingly contrived'' may signal a more negative review and lead to the item being easier.

In both cases, it is not clear if there is a ``gold standard'' for difficulty.
Estimating difficulty using IRT relies on responses from a group of humans or an ensemble of models, and the resulting difficulty estimates may be biased based on who or what provides the labels.
Human intuitions or model architecture decisions impact the response patterns collected, which in turn affect the learned parameters.
An investigation into what upstream information drives downstream effects such as learned difficulty is an interesting and important direction for future work.

\section{Related Work}

Prior work has considered IRT in the context of evaluating ML models using human \cite{lalor2016beyond} and machine-generated \cite{martinez2016making} response patterns.
\citet{martinez2016making} attempted to fit IRT models using machine generated response patterns on small data sets (i.e. 200-300 items), but obtain results that are difficult to interpret using the existing IRT assumptions.
\citet{lalor2016beyond} develop new IRT test sets for NLI using human-generated data, and present new ways to interpret and understand model performance beyond raw accuracy.
Due to the need for human annotations the resulting tests are short (i.e. 124 examples).
To the best of our knowledge no one has attempted to fit IRT models using DNN-generated response patterns on large data sets.

There have been a number of studies on modeling latent traits of data to identify a correct label, 
\cite[e.g.][]{bruce1999recognizing}. 
There has also been work in modeling individuals to identify poor annotators~\cite{hovy2013learning}, but neither jointly model the ability of individuals and data points, nor apply the resulting metrics to interpret DNN models.
Other work has modeled the probability a label is correct along with the probability of an annotator to label an item correctly according to the~\cite{dawid1979maximum} model, but do not consider difficulty or discriminatory ability of the data points \cite{passonneau2014benefits}.
In the above models an annotator's response depends on an item only through its correct label. 
IRT assumes a more sophisticated response mechanism involving both annotator qualities and item characteristics. 
The DARE model \cite{DBLP:conf/icml/BachrachGMG12} jointly estimates ability, difficulty and response using probabilistic inference. 
It was evaluated on an intelligence test of 60 multiple choice questions administered to 120 individuals.  

There are several other areas of study regarding how best to use training data that are related to this work.
Re-weighting or re-ordering training examples is a well-studied and related area of supervised learning.
Often examples are re-weighted according to some notion of difficulty, or model uncertainty~\cite{chang2017active}.
In particular, the internal uncertainty of the model is used as the basis for selecting how training examples are weighted.
However, model uncertainty depends upon the original training data the model was trained on, while here we use an external measure of uncertainty.

Curriculum learning (CL) is a training procedure where models are trained to learn simple concepts before more complex concepts are introduced~\cite{bengio_curriculum_2009}.
CL training for neural networks can improve generalization and speed up convergence.
In curriculum learning the difficulty of items is typically assigned based on heuristics of the data (e.g. the number of sides of a shape).
IRT models directly estimate difficulty from the responses of human or machine test-takers themselves instead of relying on heuristics.
Self-paced learning and the Leitner method use model performance to estimate difficulties, but are restricted to a single model's performance, not a more global notion of difficulty \cite{kumar2010self,amiri2018spotting}.

\section{Conclusion}

In this work we have described how large-scale IRT models can be trained with DNN response patterns using VI.
Learning the difficulty parameters of items and the ability parameters of DNN models allows for more nuanced interpretation of model performance and enables us to filter training data so that DNN models can be trained on less data while maintaining generalization as measured by test set performance.
IRT models with machine RPs can be fit not only for NLP data sets but also data sets in other machine learning domains such as computer vision (additional results on two computer vision data sets are included in Appendix A).

One limitation of this work is the up-front cost of generating RPs from the DNN ensemble.
However, the cost of running a large number of DNN models to generate response pattern data is significantly less than the cost of obtaining those labels from human annotators in two ways.
First, the monetary cost of asking thousands of humans to label tens or hundreds of thousands of images or sentence pairs is prohibitive.
Second, since the response patterns require that a single individual provide labels for all (or most) of the data set, each individual would need to label a huge number of items.
Each individual would most likely get bored or burned out and the quality of the labels would suffer.

That said, consider for example a large company (or research lab) that runs hundreds or thousands of experiments each day on some internal data set.
Many of the experiments would not lead to significant improvements in model performance, and the outputs from those experiments would be discarded.
With the methods proposed here those outputs can be used to learn the latent parameters of the data to focus in on what exactly is working well and what isn't with respect to the models being tested and the data used to train them.
Using the previously discarded data to learn IRT models and estimate latent difficulty and ability parameters can be used to improve a variety of tasks such as model selection, data selection, and curriculum learning strategies.

IRT models assume difficulty is a latent parameter of the items and can be estimated from response pattern data.
Difficulty is directly linked to subject ability, in contrast to heuristics such as sentence length or word rarity.
Certain items may be easy or difficult for a variety of reasons.
With the methods presented here, an interesting direction for future work is to further examine why certain examples are more difficult than others.

We have shown that it is possible to fit IRT models using RPs from DNN models.
Prior work relied on human RPs to investigate the impact of difficulty on model performance \cite{lalor2017analysis}, but it is now possible to conduct similar IRT analyses with machine RPs.
This work also opens the possibility of fitting IRT models on much larger data sets.
By removing the human bottleneck, we can use ensembles of DNN models to generate RPs for large data sets (e.g. all of SNLI or SSTB instead of a sample).
Having difficulty and ability estimates for machine learning data sets and models can lead to very interesting work around such areas as active learning, curriculum learning, and meta learning.

\section*{Acknowledgements}
We thank the anonymous reviewers for their comments and suggestions.
This work was supported in part by the HSR\&D award IIR 1I01HX001457 from the United States Department of Veterans Affairs (VA).
We also acknowledge the support of LM012817 from the National Institutes of Health. 
This work was also supported in part by the Center for Intelligent Information Retrieval. 
The contents of this paper do not represent the views of CIIR, NIH, VA, or the United States Government.

\bibliography{jlalor}
\bibliographystyle{emnlp2019/acl_natbib}

\end{document}